# MRIo3DS-Net: A Mutually Reinforcing Images to 3D Surface RNN-like framework for model-adaptation indoor 3D reconstruction

Chang Li [a], Jiao Guo [a], Yufei Zhao [b] and Yongjun Zhang [c]

[a] Key Laboratory for Geographical Process Analysis & Simulation of Hubei Province, and College of Urban and Environmental Science, Central China Normal University, Wuhan, China;

[b] Hainan Surveying and Mapping Product Quality Supervision and Inspection Station, MNR, China.

[c] School of Remote Sensing and Information Engineering, Wuhan University, China.

**Abstract:**

Currently, deep neural networks (DNNs) have made some progresses in local region feature extraction, image matching under difficult conditions and point cloud surface optimization in photogrammetry and computer vision. However, there have been being no reports on an end-to-end high-accuracy 3D reconstruction network that integrated point cloud generation (i.e., dense image matching) and 3D surface optimization. Moreover, deep-learning-based 3D reconstruction networks are often difficult to apply to new dataset tasks. Therefore, this paper is the first to propose an end-to-end framework of mutually reinforcing images to 3D surface recurrent neural network (RNN)-like (MRIo3DS-Net) for model-adaptation indoor 3D reconstruction, where multi-view dense matching and point cloud surface optimization are mutually reinforced by a RNN-like structure rather than being treated as a separate issue. The characteristics and advantages of this framework are as follows: (1) In the multi-view dense matching module (MVDMM), the model-adaptation strategy is used to fine-tune and optimize a Transformer-based multi-view dense matching DNN, so that it has the higher image feature for matching and detail expression capabilities; (2) In the point cloud surface optimization module (PCSOM), the 3D surface reconstruction network based on 3D implicit field is optimized by using model-adaptation strategy, which solves the problem of point cloud surface optimization without knowing normal vector of 3D surface. Especially, to improve and finely reconstruct 3D surfaces from point cloud, smooth loss function is proposed and added to this module; (3) The proposed MRIo3DS-Net is a RNN-like framework, which utilizes the finely optimized 3D surface obtained by PCSOM to recursively reinforce the differentiable warping for optimizing MVDMM. This refinement leads to achieving better dense matching results, and better dense matching results leads to achieving better 3D surface results recursively and mutually. Hence, model-adaptation strategy can better collaborate the differences between the two network modules, so that they complement each other to achieve the better effect; (4) To accelerate the transfer learning and training convergence from source domain to target domain, a multi-task loss function based on Bayesian uncertainty is used to adaptively adjust the weights between the two networks loss functions of MVDMM and PCSOM; (5) In this multi-task cascade network framework, any modules can be replaced by any state-of-the-art (SOTA) networks to achieve better 3D reconstruction results. We tested performance of MRIo3DS-Net in the DTU, Tanks and Temples datasets by comparison experiments and ablation experiments with SOTA networks. The results show that MRIo3DS-Net has achieved the best performance and can reconstruct various indoor scenes reliably, completely and accurately. MRIo3DS-Net can provide a new end-to-end framework for high-accuracy indoor 3D reconstruction and has certain potential for the new target domain task.

## 1. Introduction

In the digital age, 3D reconstruction technology represents a fundamental pillar of human ingenuity, fulfilling an indispensable function. The concepts of digital earth and smart earth present a prospective blueprint for the future, one that is based on precise data and advanced technology (Zhang et al., 2023b). Through 3D reconstruction, we can comprehensively acquire, analyze, and display both natural and cultural information about the Earth, thereby promoting rapid development in global environmental monitoring (Tachella et al., 2019), resource management, and disaster prevention (Nagasawa et al., 2021). The rise of digital twin cities and the metaverse has made 3D reconstruction a key technology for interaction between the virtual and real worlds (Caldarelli et al., 2023). This advancement drives the development of urban planning (Han et al., 2023), smart transportation (Guo et al., 2022), and virtual reality (Gao et al., 2023b). Against this backdrop, China's Ministry of Natural Resources proposed the "national 3D real scene" initiative. Driven by technologies such as artificial intelligence and deep learning, this initiative has achieved comprehensive perception and precise management of natural resources through high-accuracy 3D models. It promotes automated and detailed 3D reconstruction of indoor and outdoor scenes, advancing the geoinformation industry(Yin et al., 2023).

Currently, outdoor 3D reconstruction technology has made significant progress through satellite and UAV photogrammetry. However, since most human activities occur indoors, the importance of indoor 3D reconstruction is even more pronounced. Indoor 3D reconstruction is a crucial and challenging task (Zhang et al., 2022), serving as the foundation for various location-based services, including indoor positioning and navigation (Tang et al., 2024), emergency management (Sakaino, 2023), and 3D mapping (Liu et al., 2023). Current indoor 3D mapping and navigation technologies are still immature and face numerous challenges. Therefore, there is an urgent need for in-depth research on indoor environments and the development of efficient and accurate 3D reconstruction methods to meet the growing application demands.

Traditional 3D reconstruction method includes mature commercial software and algorithms. They have made it possible to achieve unsupervised 3D indoor scenes reconstruction with high accuracy and reliability using traditional methods. However, achieving high overlap and intersection angles of photos is necessary. With the continuous development of deep learning, deep-learning-based 3D reconstruction has become a trend.

Many studies (Ding et al., 2022; Gao et al., 2023a; Liu et al., 2023; Zhang et al., 2023d) focus on using various DNNs for multi-view dense matching. Some researchers employ parametric or implicit learning methods for point cloud surface optimization tasks (Li et al., 2024; Wang et al., 2024). In these studies, multiple-view dense matching and point cloud surface optimization are treated as two independent tasks. However, better quality dense matching leads to higher quality point clouds. When these refined point clouds are fed into the surface optimization network, it enhances the quality of surface reconstruction (Wang et al., 2024). This optimized point cloud surface can in turn provide finer digital surface model (DSM) for multiple-view dense matching, resulting in improved dense matching results (Li et al., 2023). Therefore, multiple-view dense matching and point cloud surface optimization complement each other rather than being isolated tasks. Additionally, DNNs for 3D reconstruction often exhibit poor performance on new datasets (Liu et al., 2023). Hence, efforts are directed towards improving the generalization of 3D DNNs using domain adaption (Ganin et al., 2016; Pinheiro et al., 2019).

In summary, DNNs have made significant advancements in image matching and point cloud surface optimization, even under challenging conditions such as feature extraction of indoor scenes, poor texture and untextured regions. However, to the best of our knowledge, there are still several areas for improvement in deep-learning-based 3D reconstruction networks:

(1) There have been being no reports on an end-to-end 3D reconstruction network that integrated point cloud generation (i.e., dense image matching) and 3D surface optimization.

(2) Multi-view dense matching and point cloud surface optimization tasks are treated independently in existing DNNs. However, sparse 3D reconstruction results can optimize dense matching results, and the optimized matching results can improve 3D reconstruction accuracy, creating a mutually reinforcing process.

(3) The quality of multi-view dense matching

1) Gaps in the point cloud surface due to failed multi-view images dense

matching;

2) Point cloud errors caused by insufficient sub-pixel matching accuracy;

(4) The quality of point cloud surface: the resulting point cloud surface exhibits strong bending due to the presence of error perturbations in the surface normal direction, which arise from the multi-view dense matching;

(5) The weak generalization of deep 3D reconstruction network models: the performance of deep 3D reconstruction network models on new datasets is poor due to their weak generalization ability.

To this end, we propose a mutually reinforcing multi-task cascade network framework with multi-view dense matching module (MVDMM) and point cloud surface optimization module (PCSOM) to automatically and intelligently reconstruct 3D indoor scenes. Additionally, our proposed framework has potential for reconstructing new target domain 3D models. Therefore, our contributions are as follows.

(1) An end-to-end mutually reinforcing images to 3D surface RNN-like, i.e., MRIo3DS-Net, framework is proposed for the first time. The framework cascades two tasks, multi-view dense matching and point cloud surface optimization, with the following advantages:

1) Achieving mutually reinforcing end-to-end high-accuracy 3D reconstruction results;

2) Adopting a MVDMM based on Transformer, which has higher image feature for matching and detail expression capabilities;

3) Using a PCSOM with 3D implicit field to optimize the point cloud surface without knowing normal vector of 3D surface;

4) Establishing a framework akin to RNN, where the two modules (i.e., MVDMM and PCSOM) mutually reinforce each other to recursively obtain dense matching results and three-dimensional surface results;

5) Being replaced by any more advanced or SOTA network to obtain better 3D reconstruction results.

(2) A multi-task loss function based on Bayesian uncertainty is designed to

adaptively adjust the weights between focal loss in MVDMM and binary cross-entropy loss and smooth loss in PCSOM without manually fine tuning the weight hyperparameter in this multi-task cascade network.

(3) The model-adaptation strategy from source domain to target domain is proposed, which not only makes the 3D reconstruction based on deep learning more universal so that it can apply to new target datasets, but also accelerates the convergence of MVDMM and PCSOM. In addition, it also promotes the continuous iteration and mutual promotion between the two modules in MRIo3DS-Net, achieving better performance than that of a single module.

## 2. Related work

### 2.1 Multi-view dense matching

#### 2.1.1 CNN-based methods

Initially, DNNs utilized convolutional neural networks to extract image features and perform matching. One example is MVSNet (Yao et al., 2018). To reduce the significant memory consumption caused by 3D CNN regularization, some optimization networks based on MVSNet have been proposed, including CasMVSNet (Gu et al., 2020), CVP-MVSNet (Yang et al., 2020), UCS-Net (Cheng et al., 2020), Vis-MVSNet (Zhang et al., 2023a), R-MVSNet (Yao et al., 2019), D2HC-RMVSNet (Yan et al., 2020), and AA-RMVSNet (Wei et al., 2021). GeoMVSNet(Zhang et al., 2023d) is a feature fusion and probabilistic volume embedding method based on geometric prior. It enhances the robustness of cost matching, ensuring stable cost matching and delivering excellent 3D reconstruction results. RA-MVSNet(Zhang et al., 2023c) introduces point-to-surface distance supervision. It constructs a distance volume using a signed distance field to better predict the point-to-surface distance, thus leveraging implicit representation. UniMVSNet (Peng et al., 2022) combines the advantages of both regression and classification. It can directly constrain the cost volume like classification methods while also achieving sub-pixel depth prediction like regression methods. Furthermore, some unsupervised and self-supervised 3D reconstruction networks, such as M3VSNet (Huang et al., 2021), U-MVSNet (Xu et al., 2021), Self-supervised-CVP-MVSNet (Yang et al., 2021), have been proposed to reduce reliance on ground-truth depth values. These approaches do not require manually labeled data and reduce the need for ground-truth depth maps to some extent. However, they heavily depend on the photo-consistency assumption among multi-view images. Furthermore, while these methods have shown some improvements in matching accuracy and completeness, they still have some gaps when compared to supervised learning networks.

**2.1.2 RNN-based methods**

RNNs, as a powerful tool for processing sequential data, have played an important role in 3D reconstruction. 3D-R2N2 (Choy et al., 2016) uses an encoder-decoder architecture to unify single-view and multi-view 3D reconstruction within a framework, requiring minimal supervision. R-MVSNet (Yao et al., 2019) employs convolutional gated recurrent units (GRU) (Cho et al., 2014) instead of 3D convolutional neural networks to sequentially regularize 2D cost maps across depth. This approach reduces memory consumption and enables high-resolution reconstruction. RED-Net (Liu and Ji, 2020) employs a recurrent encoder-decoder structure instead of stacked GRU blocks to regularize 2D cost maps. This enables large-scale, full-resolution reconstruction with higher accuracy and efficiency, while reducing memory requirements. Effi-MVSNet (Wang et al., 2022) introduces a lightweight dynamic cost volume that can be processed iteratively, thus avoiding the memory and time consumption issues associated with static cost volume. Sat-MVSNet (Gao et al., 2023a) is a successful application of deep learning methods to 3D reconstruction from optical satellite images, addressing discrepancies between RPC models and perspective geometry models. Ada-MVS (Liu et al., 2023) is an adaptive multi-view cost aggregation and lightweight recurrent regularization MVS network, particularly suited for 3D scene reconstruction from oblique aerial multi-view images.

**2.1.3 Attention-based methods**

The Transformer has been introduced into 3D reconstruction tasks with the development of attention mechanisms, resulting in excellent reconstruction results (Ding et al., 2022; Li et al., 2021; Sarlin et al., 2020). SuperGlue (Sarlin et al., 2020) incorporates an attention-based, adaptable context aggregation mechanism, allowing it to simultaneously infer underlying 3D scenes and establish feature correspondences. LoFTR (Sun et al., 2021) employs self-attention and cross-attention layers within a Transformer to derive feature descriptors from pairs of images. Initially, it forms dense pixel-wise correspondences at a coarse scale and then fine-tunes these matches at a finer scale. STTR (Li et al., 2021) substitutes cost volume construction with dense pixel matching leveraging positional information and attention, thereby easing the constraints of a fixed disparity range. TransMVSNet (Ding et al., 2022) is the first to employ a Transformer for the MVS task, utilizing self-attention and cross-attention to aggregate long-range context information both within and between images.

Currently, multi-view dense matching has developed to a highly advanced level. However, existing multi-view dense matching methods rely on the results of sparse matching. Sparse matching generates a coarse (DSM) which guides the image-based matching (Wang et al., 2024). However, sparse matching often lacks accuracy, leading to inaccurate initial DSM and affecting the dense matching results and quality. Therefore, precise dense matching should consider the optimization of the DSM, ensuring an accurate surface approximation(Li et al., 2023). This is an iterative process. Thus, building on existing DNN-based multi-view dense matching work, we propose, for the first time, a dense matching network based on a Transformer that iteratively optimizes the DSM.

## 2.2 Point cloud surface optimization

Deep-learning-based point cloud surface optimization methods have evolved with the deepening of neural networks and can be roughly categorized into parametric methods and implicit learning methods. AtlasNet (Groueix et al., 2018) was initially proposed for representing closed surfaces only. Williams et al. (2019) also proposed a parametric network for surface reconstruction. This algorithm can effectively robustly fit noisy inputs and capture sharp features of objects, but it is computationally very slow. Recently, point cloud surface optimization based on implicit representation has seen rapid development, enabling high-resolution reconstruction results. DeepSDF (Park et al., 2019) directly obtains a continuous signed distance function from a noisy point cloud through an auto-decoder, generating watertight surfaces for objects with complex topologies. However, it cannot achieve high-quality surface reconstruction for large-scale point clouds. Subsequently, ONet (Mescheder et al., 2019) was introduced, which can directly reconstruct objects using a continuous 3D voxel occupancy grid. However, it is comparable to DeepSDF (Park et al., 2019). Therefore, it cannot generate surfaces for large-scale point clouds. Atzmon et al. (2020) introduced a variational auto-encoder into their algorithm to address the issue of reconstructing high-fidelity surfaces from the original point cloud. The auto-encoder directly learns an implicit shape representation from the raw point cloud. However, it is still unable to reconstruct the surfaces of thin structures. POCO (Boulch and Marlet, 2022) employs point cloud convolution to compute latent vectors for each point, and uses inferred weights to perform learned interpolation on the nearest neighbors. ALTO (Wang et al., 2023b) transitions sequentially between geometric representations before settling into a latent function that is simple to decode. This method preserves spatial expressiveness and keeps decoding lightweight. GridFormer (Li et al., 2024) introduces a boundary optimization strategy that combines boundary binary cross-entropy loss and boundary sampling, enabling more precise representation of object structures. Sp-ConvONet (Wang et al., 2024) introduces a neural implicit method based on sparse convolution, focusing feature and network computations only on grid points close to the surface to be reconstructed. This enables reconstruction of higher-resolution 3D grids with high-fidelity details.

Currently, point cloud surface optimization has reached a mature stage, yielding impressive results. However, existing methods treat point cloud surface optimization as a separate, independent module. In reality, it is closely related to the dense matching module. The quality of the input point cloud directly impacts the result of point cloud surface optimization. Better quality dense matching produces higher quality point clouds, which, when fed into the surface optimization network, can enhance the quality of surface optimization. Building on the implicit, normal vector-independent point cloud surface optimization networks, we propose, for the first time, a combined dense matching and point cloud surface iterative optimization module while maintaining surface smoothness. The output surface guides high-quality dense matching. Better dense matching quality generates superior point clouds, which are fed into the surface optimization network, resulting in higher-quality surface. This creates a positive feedback loop.

**2.3 Domain adaption**

In domain-adaptive 3D reconstruction, researchers employ various techniques such as adversarial learning to narrow the gap between the source domain and the target domain. Adversarial learning, through the interplay between generator and discriminator, progressively reduces the distribution gap between the source and target domains, thereby enhancing the performance on the target domain. Ganin et al. (2016) addresses this issue by employing an adversarial objective that forces domain confusion between two domains. Pinheiro et al. (2019) proposed a framework for domain-adaptive 3D reconstruction using adversarial learning. They apply domain confusion between natural and synthetic image representations to reduce distribution gap. Additionally, they enforce the reconstruction to adhere to the manifold of real-world object shapes, enhancing the realism of the reconstructed results.

Adversarial learning techniques present a promising strategy for training robust deep networks, capable of generating complex samples across different domains. However, these methods often involve complex models and face challenges with convergence. Therefore, exploring ways to fine-tune models to enhance network generalization without increasing model complexity and ensuring easier convergence is worthwhile. To this end, we adopted a model adaptation strategy, enhancing model generalization by freezing and fine-tuning the network.

**2.4 Multi-task learning**

Traditional 3D reconstruction methods typically handle tasks like multi-view dense matching and point cloud surface optimization separately. However, this single-task approach often struggles to fully leverage the correlated information between different tasks. In research utilizing multi-task learning for 3D reconstruction, many researchers have proposed various methods to effectively leverage the correlations between tasks. Kendall et al. (2018) significantly improved the accuracy and consistency of 3D reconstruction by integrating depth estimation, semantic segmentation, and instance segmentation tasks through the introduction of a joint loss function. Wang et al. (2023a) proposed an uncertain multi-task optimization method to adaptively combine matching metrics and semantic metrics. Huang et al. (2024) utilize multi-view high-level semantic information to maintain semantic consistency, aiding in the optimization of MVS networks.

Despite significant advancements in multi-task learning for 3D reconstruction, challenges remain in effectively balancing the weights between multi-view dense matching and point cloud surface optimization. Unlike existing dense matching and point cloud surface optimization networks, we consider multiple tasks for the first time, including multi-view dense matching, point cloud surface optimization, point cloud surface smooth constraints, and recurrent networks, for comprehensive optimization.

Above all, to the best of our knowledge, there have been being no reports on end-to-end high-accuracy 3D reconstruction networks integrating point cloud generation and 3D surface optimization based on deep learning. Moreover, deep-learning-based 3D reconstruction networks often struggle to adapt to new datasets and tasks. To address these issues, we propose an end-to-end framework for 3D indoor scene reconstruction. It has the following characteristics: (1) within this framework, refined 3D surfaces provide feedback to improve differentiable warping accuracy, and the improved differentiable warping results further refine 3D surface optimization, forming RNN-like mode; (2) the framework employs model-adaptive strategies to address weak generalization issues in deep-learning-based 3D reconstruction networks; (3) a multi-task loss function adaptively adjusts the weights between focal loss in MVDMM, and binary cross-entropy and smooth loss in PCSOM, eliminating the need for manual hyperparameter tuning in this multi-task cascade network. The mutual reinforcement among modules within this multi-task cascade network framework enhances the automation, intelligence, and integration level of 3D reconstruction.

## 3 Methodology

### 3.1 Overview

The MRIo3DS-Net is a mutually reinforcing multi-task cascade network where MVDMM and PCSOM are connected in series. This setup allows them to recursively or mutually reinforce each other mutually. Firstly, multi-view images are processed using mature automatic aerial triangulation, including sparse image matching and self-calibration bundle adjustment. Secondly, the results serve as input to MRIo3DS-Net framework. The proposed MRIo3DS-Net framework includes a multi-view dense matching network based on Transformer and model-adaptation strategy to interconnect with new dataset tasks (i.e., MVDMM) and an 3D surface optimization network based on 3D implicit field (i.e., PCSOM), which can not only interconnect with new dataset tasks but also have faster convergence speed after model adaptation. Specially, after the result of MVDMM is inputted into PCSOM, the result of PCSOM is also inputted into MVDMM, thus forming a recurrent optimization. Finally, a multi-task loss function is designed based on Bayesian uncertainty to adjust the weight of MVDMM and PCSOM.

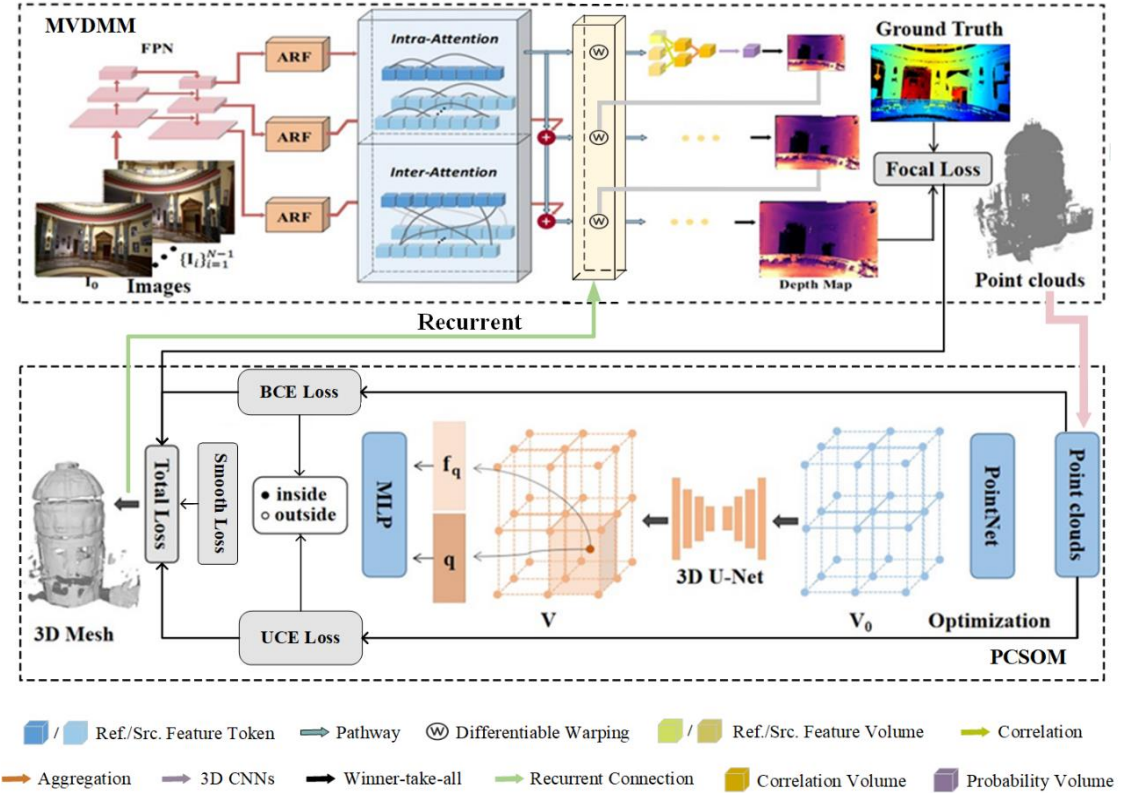

Figure 1. The proposed MRIo3DS-Net structure

Figure 1 displays the structure of MRIo3DS-Net framework, whose steps are as follows:

(1) Automatic aerial triangulation: including sparse matching and self-calibration bundle adjustment. The results of this step are used as input for the MRIo3DS-Net framework;

(2) MVDMM: taking multi-view images and sparse point clouds as inputs and using multi-view dense matching network based on Transformer to generate dense point cloud;

(3) Model adaptation on multi-view dense matching: transferring MVDMM from the original dataset to the target dataset to accelerate the convergence of MRIo3DS-Net framework based on the model-adaptation strategy;

(4) PCSOM: finishing 3D reconstruction by means of an implicit 3D surface optimization network, with dense point cloud as input

(5) Model adaptation on point cloud surface optimization: using model-adaptation strategy to transfer PCSOM from the original dataset to the target dataset and accelerate the convergence of MRIo3DS-Net framework.

(6) RNN-like inter-module reinforcement: acquiring the finely optimized 3D surface from PCSOM, which provides feedback to the differentiable warping of MVDMM, aiming to achieve more accurate dense matching results and point clouds, forming a RNN-like inter-module mutually reinforcing recursive optimization strategy；

(7) Multi-task cascade network optimization: designing a multi-task loss function to automatically adjust the weight of MVDMM and PCSOM, improve the accuracy and robustness of the network, and further optimizing the visualization effect of 3D reconstruction.

## 3.2 Multi-view dense matching module

As one of the choices for the MVDMM of MRIo3DS-Net framework, TransMVSNet(Ding et al., 2022) analogizes MVS back to its nature of a feature matching task, and introduces Transformer into MVS and proposes a Feature

Matching Transformer (FMT, Fig. 2), which follows (Sarlin et al., 2020) by adding position encoding to enhance position consistency and robustness to feature maps with different resolutions. Then the feature is input into the attention block where the reference feature and source feature first compute the intra-attention respectively, and then compute the unidirectional inter-attention to aggregate the context information within and between the images.

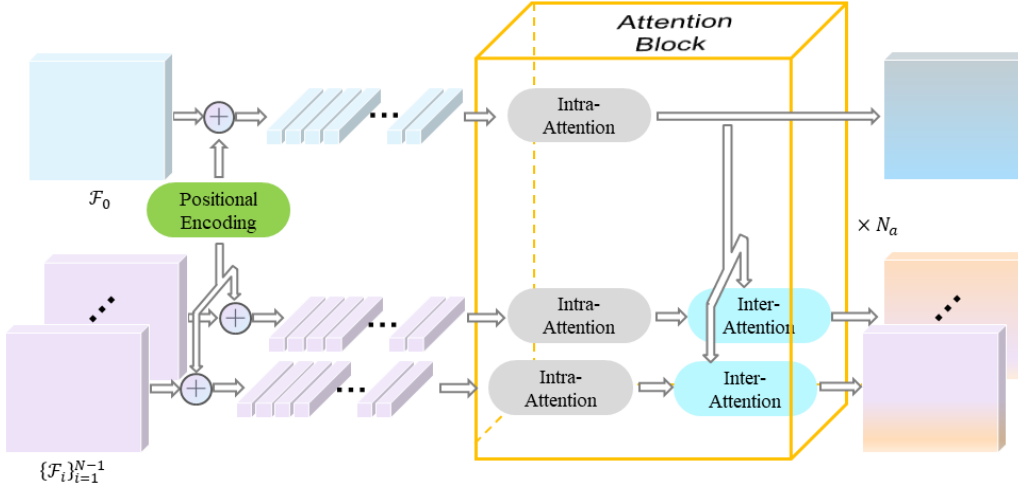

Figure 2 Architecture of the Feature Matching Transformer (Ding et al., 2022)

The attention mechanism can be characterized as the process of associating a query **Q** with a series of key (**K**)-value (**V**) pairs to generate an output, where all components, including the query, keys, values, and output, are represented as vectors (Vaswani et al., 2017). Based on attention weights derived from the dot product of **Q** and **K**, **Q** retrieves relevant information from **V**. The output of attention layer is denoted as:

$$\text{Attention}(\mathbf{Q},\mathbf{K},\mathbf{V}) = \text{softmax}(\mathbf{Q}\mathbf{K}^{\text{T}})\mathbf{V} \qquad (1)$$

TransMVSNet utilizes FPN (Lin et al., 2017b) as the feature extractor and Transformer as the encoder. To ensure a smooth transition of the range of features, an Adaptive receptive field module is inserted between FPN and FMT. Deformable convolution (Dai et al., 2017; Zhu et al., 2019) is utilized to learn additional offsets for sampling positions, which adaptively expands the receptive fields. Then the features of different image scales are bridged by a transformed feature pathway to transfer the transformed features and gradients across different image scales.

Additionally, the network uses differentiable warping to align all images with the reference image and pair-wise feature correlation to measure feature similarity.

The warping of pixel $p$ in the reference image and its corresponding pixel $\hat{p}$ in the source image under depth estimation $d$ is defined as:

$$\hat{p}=\mathbf{K}[\mathbf{R}(\mathbf{K}_0^{-1}pd)+t] \tag{2}$$

where $\mathbf{K}_0$ and $\mathbf{K}$ are the intrinsic matrices of the reference and the target camera respectively. $\mathbf{R}$ and $t$ are the rotation and translation transformations between the two views.

In addition, to improve the photo-consistency between images at pixel $x$, we further constraints the reprojection error with Photo-Consistency Refinement loss (Faugeras and Keriven, 2002; Pons et al., 2007; Vu et al., 2012):

$$L(S) = \sum_{i,j} \int_{\Omega_{ij}^S} -h(I_i, I_{ij}^S)(x_i) \mathbf{N}^\mathrm{T} \mathbf{d}_i \mathrm{d}x/z_i^3 \tag{3}$$

where $x_i = \prod_i(x)$, $x_j = \prod_j(x)$; $h(I, J)(x_i)$ aims to measure the photo-consistency between images $I$ and $J$ at pixel $x$; $I_{ij}^S = I_j \circ \prod_j \circ \prod_i^{-1}$ is the reprojection from image $I_j$ to $I_i$ induced by the object surface $S$; $\Omega_{ij}^S$ is the domain of this reprojection. $\prod_i$ and $\prod_i^{-1}$ are the projection and inverse projection from image $i$ to $S$ respectively defined by Eq.2. $\mathbf{N}$ is the outward surface normal at $x$; $\mathbf{d}_i$ is the vector joining the center of camera $i$. $z_i$ is the depth of $x$ in camera $i$.

Correlation volume is defined as:

$$h(I_i, I_{ij}^S)^{(d)}(\boldsymbol{p}) = \sum_{i=1}^{N-1} \max_d \left\{ h_i^d(\boldsymbol{p}) \right\} \cdot h_i^d(\boldsymbol{p}) \tag{4}$$

where $h_i^{(d)}(\boldsymbol{p})$ is the correlation volume at position $\boldsymbol{p}$, $h_i^{(d)}(\boldsymbol{p}) = <\mathcal{F}_0(\boldsymbol{p}), \hat{\mathcal{F}}_i^{(d)}(\boldsymbol{p})>$, where $\hat{\mathcal{F}}_i^{(d)}(\boldsymbol{p})$ represents the warped $i$-th source feature map at depth $d$.

The network treats depth estimation as a classification task and uses focal loss (Lin et al., 2017a) to reduce ambiguity and strengthen supervision:

$$L_F = \sum_{\boldsymbol{p} \in \{p_v\}} -\left(1 - P^{(\bar{d})}(\boldsymbol{p})\right)^\gamma \log\left(P^{(\bar{d})}(\boldsymbol{p})\right) \tag{5}$$

where $P^{(\bar{d})}(p)$ represents predicted probability of the depth hypothesis $\bar{d}$ at pixel $p$, and $\{p_v\}$ represents the subset of pixels with a valid ground truth. When the focusing parameter $\gamma$ equals 0, focal loss degenerates to cross entropy loss.

**3.3 Point cloud surface optimization module**

As one of the choices for the PCSOM of MRIo3DS-Net framework (Fig. 3), SA-ConvONet (Tang et al., 2021) approximates a surface by predicting a neural implicit field **O**. First, a shallow PointNet (Qi et al., 2017) is used to process the input point cloud to obtain point-wise features. The point features belonging to the same voxel cell are integrated using average pooling. The volumetric features $V$ are obtained by aggregating global and local information using 3D U-Net.

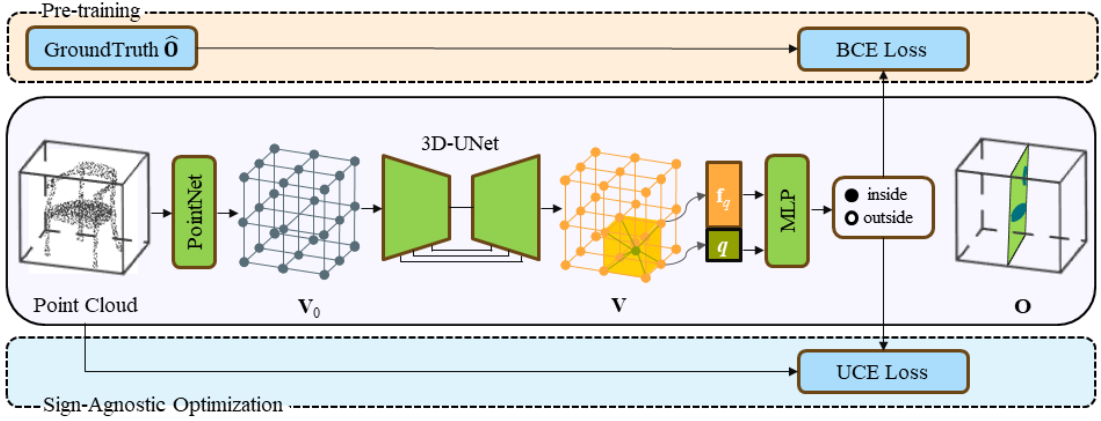

Figure 3 SA-ConvONet Overview (Tang et al., 2021)

Then, SA-ConvONet can obtain the occupancy probability $O(q) \in \mathbb{R}^3$ of randomly sampled point $q$ in 3D space based on $V$. To achieve this, the feature vector $f_q$ corresponding to $q$ is queried in $V$ by trilinear interpolation. The resulting feature vector is then input into the occupancy decoder operator $D$, which is constructed using a light-weight multi-layer perceptron network:

$$O(q) = \left(1 + e^{-D(q, \mathbf{f}_q)}\right)^{-1} \in (0,1) \qquad (6)$$

During training, the paper uses binary cross-entropy loss between the predicted and the true occupancy values as a loss function:

$$L_B = \sum_{q \in Q} \left[ -\hat{O}(q) \log O(q) - \left(1 - \hat{O}(q)\right) \log\left(1 - O(q)\right) \right] \qquad (7)$$

where $O(q)$ is the predicted occupancy value, $\hat{O}(q)$ is the true occupancy value.

In addition, SA-ConvONet further optimizes the pre-trained model for the given input if the given input deviates significantly of pre-trained priors. Specifically, in the absence of a surface normals, SA-ConvONet uses unsigned cross entropy loss to obtain consistency constraints between the occupancy field and the unsigned input. The predicted occupancy value O($q$) and target $\hat{O}(q)$ are defined as follows:

$$O(q) = \left(1 + e^{-|D(q, f_q)|}\right)^{-1} \in [0.5, 1) \tag{8}$$

$$\hat{O}(q) = \begin{cases} 0.5, q \in S \\ 1.0, q \in \hat{S} \end{cases} \tag{9}$$

Where $P_S$ is the set of points obtained from the ground truth surface $S$, $P_{\hat{S}}$ is the set of points sampled from a non-surface volume $\hat{S}$.

Additionally, we further propose to add a smooth loss (Vu et al., 2011) to impose smooth constraints on the reconstructed point cloud surface:

$$L_S = \int_S \left(K_1^2 + K_2^2\right) dS \tag{10}$$

where $K_1$ and $K_2$ are the principal curvatures of a point on the surface $S$. In practical applications, this paper employs the discretized form of Eq. 10 to calculate the smooth loss.

The loss function of this module is defined as the weighted sum of the above two loss functions.

**3.4 RNN-like inter-module reinforcement**

The network uses multiresolution IsoSurface extraction (Mescheder et al., 2019) and Marching Cubes (Lorensen and Cline, 1987) to extract the surface meshes as the final reconstruction results. The MRIo3DS-Net framework integrates the MVDMM and PCSOM, mutually reinforcing them instead of treating them as separate issues (Fig. 4). The PCSOM generates a finely optimized 3D surface, which is feedback to MVDMM to optimize its differentiable warping. This results in more accurate multi-view dense matching and more precise point clouds. These are then inputted into PCSOM for point cloud surface optimization, creating a RNN-like inter-module mutually reinforcing optimization strategy:

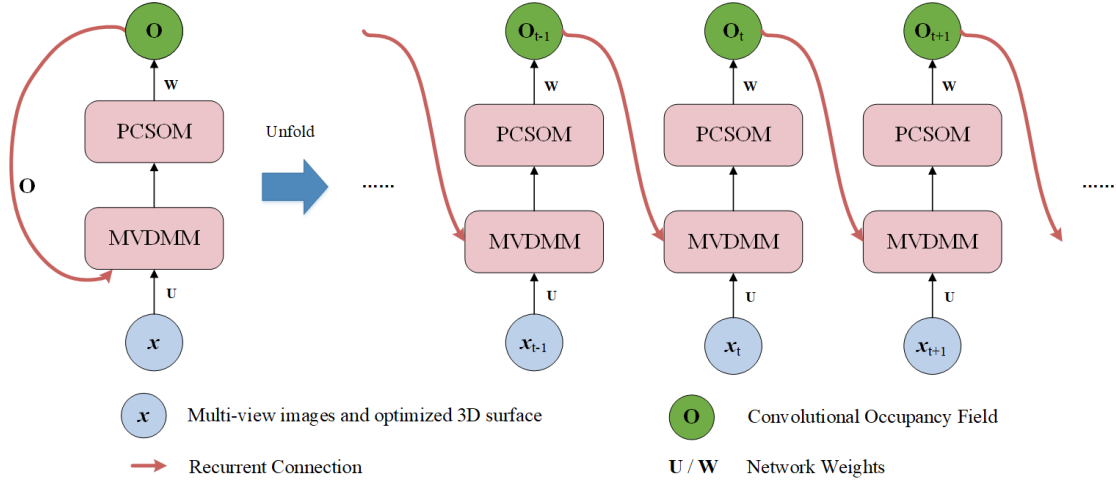

Figure 4 RNN-like inter-module (i.e., mutual) reinforcement

$$h_t = f(\mathbf{U}_{hx}x_t + \mathbf{W}_{hh}h_{t-1} + b_h) \tag{11}$$

$$\hat{\mathbf{O}}_t = \text{sigmoid}(\mathbf{W}_{hO}h_t + b_O) \tag{12}$$

where $x_t$ is the input of RNN at the *t*-th iteration, i.e., the multi-view images and optimized 3D surface; $\hat{\mathbf{O}}_t$ is the output at the *t*-th iteration. $h_t \in \mathbb{R}^N$ is the hidden state at the *t*-th iteration, i.e., the MVDMM and PCSOM. $\mathbf{U}_{hx}$ is the weight matrix from the input to the hidden state. $\mathbf{W}_{hh}$ is the weight matrix from the previous hidden state to the current hidden state. $\mathbf{W}_{hO}$ is the weight matrix from the hidden state to the output. $b_h$, $b_O$ are the bias for the hidden layer and the output layer, respectively.

### 3.5 Model Adaptation

To address the challenge of applying the deep-learning-based 3D reconstruction network to new dataset tasks and to expedite convergence of the multi-task cascade MRIo3DS-Net framework, this paper conducts model adaptation on the new datasets of MVDMM and PCSOM respectively. Assuming that *X* is the input space and *Y* is the output space, domain adaptation is the ability to apply algorithm trained in one or more source domains $D_S$ to different but related target domain $D_T$ at $X \times Y$ (Sun et al., 2015). Model adaptation is a kind of domain adaptation. Its key idea is to train the pre-trained model with labeled or unlabeled samples of the target domain $D_T$, and reduce the errors made in $D_T$ by learning the information of the $D_T$. Unlike fine-tune, model adaptation tends to introduce additional mechanisms into the network to achieve model adaptation (Fig. 5), such as fixing the weights and structure, i.e., freeze some layers in network, changing or adding others to train the model, etc.

Assuming the model is *f*, model adaptation can be performed by minimizing a specific loss function:

$$L = L_S + \lambda L_{adapt} \tag{13}$$

where $L_S$ is the loss function of the source domain. $\lambda$ is a hyperparameter that balances the two losses. $L_{adapt}$ is the loss function of domain adaptation that measures the difference between the source domain and the target domain, i.e., maximum mean discrepancy (MMD), $L_{adapt}=\text{MMD}(\phi(D_s),\phi(D_t))$, $\varphi$ is the feature extractor of the model. The key point of model adaptation is to optimize the overall loss function (Eq. 13), so that the model can adapt to the target domain while performing well in the source domain.

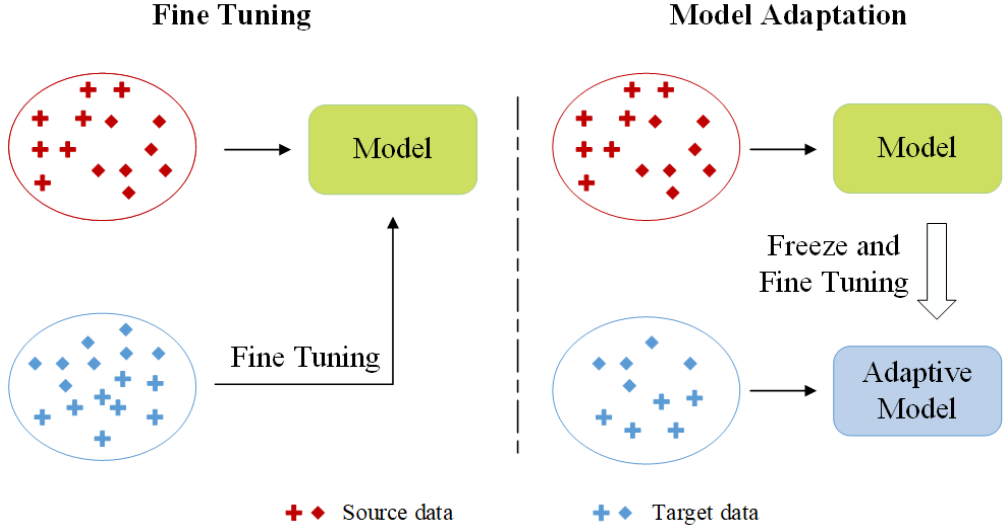

Figure 5 Model Adaptation

This paper employs a model-adaptation strategy to enhance the robustness and portability of the network on a new dataset. Specifically, it freezes some layers except the last two in the pre-trained model and fine-tunes this network. Additionally, it addresses the challenge of applying deep-learning-based 3D reconstruction networks to new dataset tasks.

### 3.6 Multi-task loss function

This paper proposes a new loss function for the multi-task cascade network inspired by the homoscedastic uncertainty between sub-tasks. The loss function balances the weightings of different sub-tasks and adjusts the loss of different image scales simultaneously to improve the network's prediction accuracy.

In general, the loss function of multi-task can be defined as a weighted sum of losses of two tasks:

$$L_{total} = \alpha \cdot L_F + \beta \cdot L_B + \delta L_S + \frac{1}{\alpha} + \frac{1}{\beta} + \frac{1}{\delta} \tag{14}$$

where $L_F$ and $L_B$, $L_S$ describe the loss functions of MVDMM and PCSOM respectively, $\alpha, \beta, \delta$ are hyperparameters. The final loss is defined as:

$$\begin{aligned} L_{total} = & \alpha \cdot \sum_{q \in Q} -\left(1 - P^{(\tilde{d})}(q)\right)^{\gamma} \log\left(P^{(\tilde{d})}(q)\right) + \\ & \beta \cdot \sum_{q \in Q} \left[-\hat{O}(q) \log O(q) - \left(1 - \hat{O}(q)\right) \log(1 - O(q))\right] \\ & + \delta \cdot \sum_{i=1}^{N} \left(\kappa_{1i}^2 + \kappa_{2i}^2\right) + \frac{1}{\alpha} + \frac{1}{\beta} + \frac{1}{\delta} \end{aligned} \tag{15}$$

However, the fine-tuning of $\alpha$, $\beta$, $\delta$ is costly, and the model is particularly sensitive to them, resulting in low accuracy of the model. Therefore, a multi-task loss (Kendall et al., 2018) based on maximizing the Gaussian likelihood with homoscedastic uncertainty is adopted in this paper to adaptively adjust the weight of two sub-tasks. Assuming that the outputs of the two modules are $y_1$ and $y_2$, $y_3$ respectively, the multi-task likelihood function is as follows.

$$p(y_1, y_2, y_3 | f^{\mathbf{W}}(x)) = p(y_1 | f^{\mathbf{W}}(x)) \cdot p(y_2 | f^{\mathbf{W}}(x)) \cdot p(y_3 | f^{\mathbf{W}}(x)) \tag{16}$$

where $f^{\mathbf{W}}(x)$ is the output of a neural network whose input is $x$ and weighting are $\mathbf{W}$, $p$ is the likelihood function.

In this case, maximizing the log likelihood of the model (Eq.16) is equal to minimizing the following objective loss function:

$$\begin{aligned} L_{total}(\mathbf{W}, \sigma_1, \sigma_2, \sigma_3) &= -\log p(y_1, y_2, y_3 | f^{\mathbf{W}}(x)) \\ &\propto \frac{1}{2\sigma_1^2} \|y_1 - f^{\mathbf{W}}(x)\|^2 + \frac{1}{2\sigma_2^2} \|y_2 - f^{\mathbf{W}}(x)\|^2 + \frac{1}{2\sigma_3^2} \|y_3 - f^{\mathbf{W}}(x)\|^2 \\ &\quad + \log \sigma_1 \sigma_2 \sigma_3 \\ &= \frac{1}{2\sigma_1^2} L_F(\mathbf{W}) + \frac{1}{2\sigma_2^2} L_B(\mathbf{W}) + \frac{1}{2\sigma_3^2} L_S(\mathbf{W}) \\ &\quad + \log \sigma_1 + \log \sigma_2 + \log \sigma_3 \end{aligned} \tag{17}$$

where $\sigma_1, \sigma_2, \sigma_3$ are observation noise parameters that affect the relative weight of loss function $L_F(\mathbf{W})$, $L_B(\mathbf{W})$ and $L_S(\mathbf{W})$ of multi-view dense matching and point cloud surface optimization, respectively. when $\sigma_1$, $\sigma_2$ and $\sigma_3$ increase, the weight of the corresponding loss functions $L_F$, $L_B$ and $L_S$ decrease. The network makes the loss approach to 0 as much as possible, so that the two noise parameters increase too much and the data is completely ignored. Therefore, we add the last term in the objective loss function as a regulariser to the loss function.

**3.7 Evaluation metrics**

We adopt the following metrics to evaluate the quality of the 3D reconstruction result.

(1) Chamfer Distance (CD): This metric measures the distance between two point clouds.

$$\mathrm{CD}(P,T) = \frac{1}{|P|}\sum_{p \in P}\min_{p' \in T}\|\boldsymbol{p} - \boldsymbol{p'}\|^2 + \frac{1}{|T|}\sum_{p' \in T}\min_{p \in P}\|\boldsymbol{p'} - \boldsymbol{p}\|^2 \tag{18}$$

where $P$ and $T$ are the set of points on the reconstructed surface and the ground truth surface, respectively. $|P|$ and $|T|$ are the number of points in the set $P$ and $T$, respectively. $\boldsymbol{p}$ and $\boldsymbol{p'}$ are points on $P$ and $T$, respectively.

(2) Normal Consistency (NC): This metric measures the average cosine of the angle between the normal vectors of the reconstructed surface and the ground truth surface, emphasizing the alignment of surface normals.

$$\mathrm{NC} = \frac{1}{|P|}\sum_{p \in P}|\boldsymbol{n}_p \cdot \boldsymbol{n}_{p'}| \tag{19}$$

where P is the set of points on the reconstructed surface; $\boldsymbol{n}_p$ is the normal vector at point $\boldsymbol{p}$ on the reconstructed surface; $\boldsymbol{n}_{p'}$ is the normal vector at the corresponding point on the ground truth surface; $|P|$ is the number of points in the set $P$.

(3) *F*-score (FS): In this paper, the *F*-score is reported with thresholds of 0.01.

(4) Accuracy: The mean absolute point-cloud-to-point-cloud distance from the reconstructed surface *P* to ground truth surface *T*.

$$\text{Acc} = \frac{1}{|P|} \sum_{p \in P} \min_{p' \in T} \| p - p' \|_2 \tag{20}$$

(5) Completeness: The mean absolute point-cloud-to-point-cloud distance from the ground truth surface *T* to the reconstructed surface *P*.

$$\text{Comp} = \frac{1}{|T|} \sum_{p' \in T} \min_{p \in P} \| p' - p \|_2 \tag{21}$$

(6) Overall: The average of Accuracy and Completeness, which indicates the overall performance of models.

## 4 Experiment results and analysis

In this section, we present the dataset and training details of the experiment. The qualitative and quantitative experimental results on different datasets are given. First, in order to verify the performance of the MVDMM and PCSOM in this paper respectively, we conduct comparison experiments between the classical networks and the DNNs in this paper, respectively in sections 4.3 and 4.4. Then, in section 4.5, comparison experiments between the MRIo3DS-Net framework and some advanced 3D reconstruction networks are presented. In section 4.6, we design experiments to prove the effectiveness of the designed multi-task loss function. Finally, the ablation experiment of MRIo3DS-Net framework is presented in section 4.7.

### 4.1 Datasets

We train the MRIo3DS-Net on DTU dataset (Aanæs et al., 2016) and Tanks and Temples (Knapitsch et al., 2017). Due to the differing datasets for the two subtasks, we optimized the aforementioned datasets before training. Specifically, we applied the poisson surface reconstruction by ContextCapture to process and optimize the point cloud in DTU dataset and Tanks and Temples, resulting in optimized point cloud surfaces. In addition, DTU dataset and Tanks and Temples are used for the training and testing of deep-learning-based multi-view dense matching. ShapeNet dataset (Chang et al., 2015) and synthetic indoor scene dataset (Peng et al., 2020) are used to perform object-level point cloud surface optimization.

## 4.2 Implementation Details

We implement MRIo3DS-Net with PyTorch, on Ubuntu16.04, with 2 NVIDIA GeForce RTX 3060 GPUs (12GB memory of each GPU) and Intel i9-10900X CPU (64GB memory).

At training phase, we set the number of views 5 for each reference view. The model is trained with a total of 10 epochs and an initial learning rate of 0.001, which decays exponentially with the number of iterations. The parameter settings of the model in the training phase are shown in Tab. 1.

Table 1 Main parameters of the model in training phase

| Settings | Patch Size | Learning Rate | Decay | Epoch | Step |
|---|---|---|---|---|---|
| MRIo3DS-Net | 512×640/576×768 | 0.001 | 0.25/0.5 | 10 | 10000 |

## 4.3 Deep-learning-based multi-view dense matching

In this paper, we first train the network on DTU dataset. In the test phase, we select evaluation set of DTU and advanced datasets of Tanks and Temples benchmark to evaluate the effect of 3D reconstruction network and the robustness of the model.

### 4.3.1 Evaluation on DTU dataset

We evaluate the proposed method on the evaluation set of DTU dataset. The results of each algorithm were compared and analyzed by three evaluation metrics: Accuracy, Completeness and Overall. Specifically, we select four representative scenes of rich texture, object occlusion, repetitive patterns and non-Lambertian surfaces for experiments, so as to demonstrate the 3D reconstruction results of all algorithms. Quantitative comparisons are shown in Tab. 2.

Table 2 Quantitative evaluation on DTU dataset (in millimeters)

| Method | Accuracy | Completeness | Overall |
|---|---|---|---|
| Gipuma(Galliani et al., 2015) | 0.283 | 0.873 | 0.578 |
| COLMAP(Schonberger and Frahm, 2016) | 0.400 | 0.664 | 0.532 |
| MVSNet(Yao et al., 2018) | 0.456 | 0.646 | 0.551 |
| CasMVSNet(Gu et al., 2020) | 0.325 | 0.385 | 0.355 |
| Uni-MVSNet(Peng et al., 2022) | 0.352 | 0.278 | 0.315 |
| TransMVSNet(Ding et al., 2022) | 0.321 | 0.289 | 0.305 |
| GBi-Net(Mi et al., 2022) | 0.327 | 0.268 | 0.298 |
| RA-MVSNet(Zhang et al., 2023c) | 0.326 | 0.268 | 0.297 |
| GeoMVSNet(Zhang et al., 2023d) | 0.331 | 0.259 | 0.295 |
| MRIo3DS-Net (ours) | **0.274** | **0.246** | **0.260** |

In Table 2, the bold values indicate the optimal 3D reconstruction results (lower the better). Compared to all known methods of traditional and DNN multi-view 3D reconstruction algorithm, the proposed method achieves better performance.

We also show quantitative visualizations of our approach and the baselines on DTU dataset (Fig. 6).

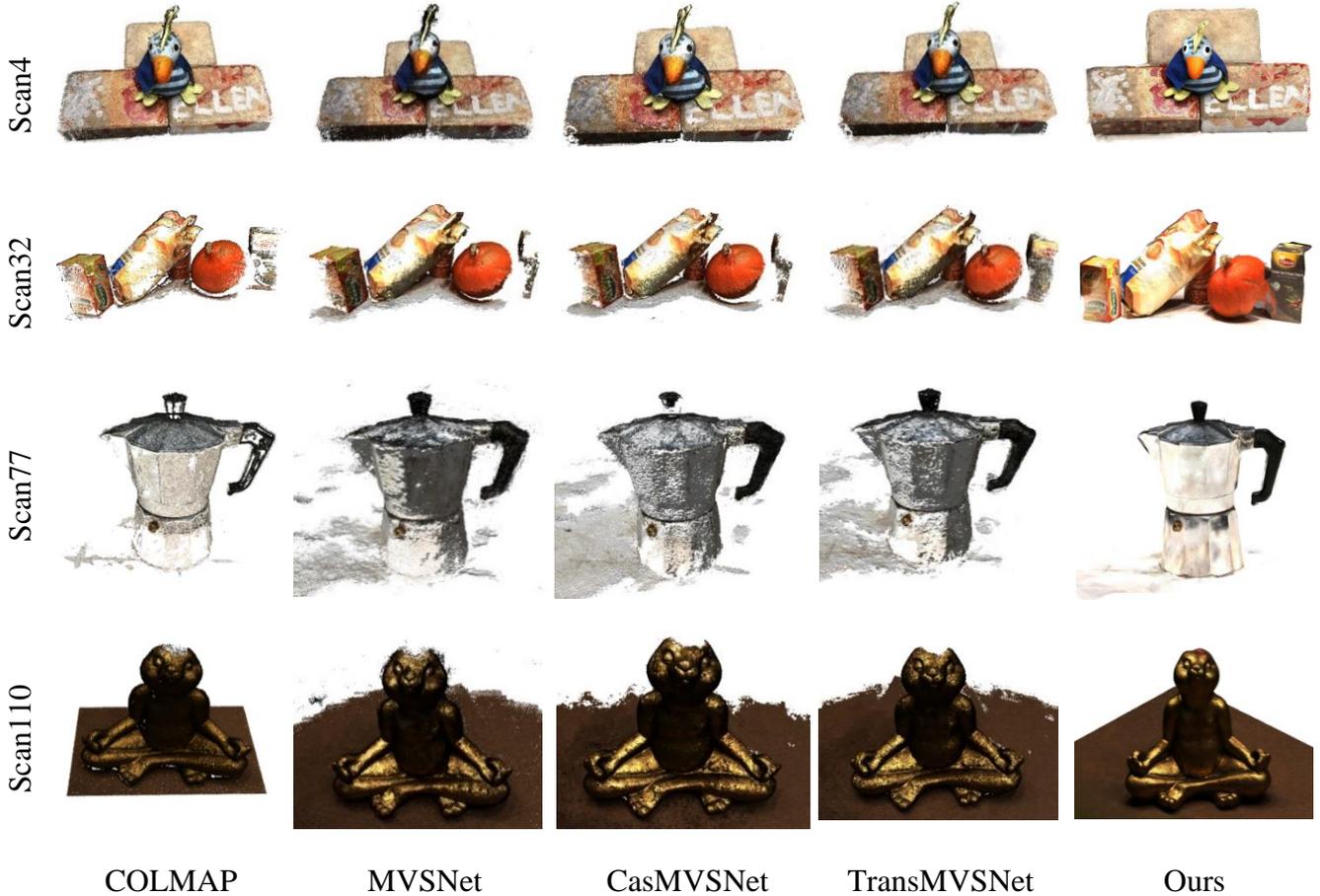

COLMAP    MVSNet    CasMVSNet    TransMVSNet    Ours

Figure 6 Quantitative results with baselines on DTU dataset.

According to Figure 6, both traditional algorithm and DNNs can complete high-quality 3D reconstruction in scenes with rich textures, but the point cloud reconstructed by COLMAP is relatively sparse and incomplete, especially in Scan77 scenes. The results of MVSNet have many noises and the edge is not smooth enough. Both CasMVSNet and TransMVSNet are designed using cascade network and feature pyramid network to better aggregate global context and generate more dense point clouds. Our framework performs more accurate and complete 3D reconstruction of poor texture, repetitive patterns or non-Lambertian surfaces in the scenes.

### 4.3.2 Evaluation on Tanks and Temples

In addition, we conducted comparative experiments with SOTA methods and our method on four large indoor scenes in Tanks and Temples for comparison and analysis. We consider mean F-score as the evaluation metrics to measure the comprehensive performance of the networks. The larger the mean F-score value is, the more dense, accurate and complete the point cloud generated by the network is.

Table 3 Quantitative evaluation results on Tanks and Temples

| Method | Auditorium | Ballroom | Courtroom | Museum | mean FS |
| --- | --- | --- | --- | --- | --- |
| COLMAP (Schonberger and Frahm, 2016) | 16.02 | 25.23 | 34.70 | 41.51 | 29.36 |
| CasMVSNet (Gu et al., 2020) | 19.81 | 38.46 | 29.10 | 43.87 | 31.12 |
| Effi-MVSNet(Wang et al., 2022) | 20.22 | 42.39 | 33.73 | 45.08 | 34.39 |
| GBi-Net(Mi et al., 2022) | 29.77 | 42.12 | 36.30 | 47.69 | 37.32 |
| TransMVSNet (Ding et al., 2022) | 24.84 | 44.59 | 34.77 | 46.49 | 37.00 |
| Uni-MVSNet(Peng et al., 2022) | 28.33 | 44.36 | 39.74 | 52.89 | 38.96 |
| RA-MVSNet (Zhang et al., 2023c) | 29.17 | 46.05 | 40.23 | 53.22 | 39.93 |
| GeoMVSNet (Zhang et al., 2023d) | **30.23** | 46.53 | 39.98 | 53.05 | 41.52 |
| MRIo3DS-Net (ours) | 29.73 | **49.61** | **42.15** | **54.84** | **44.42** |

The quantitative results are illustrated in Tab. 3. Compared with other networks, Ours' achieves impressive results, demonstrating its capability of yielding high-quality 3D reconstruction. This is mainly due to the model-adaptation strategy and the fact that TransMVSNet adopts FPN to extract multi-scale feature maps. In addition, the self- and cross-attention of the feature matching can also better focus edge and ambiguous areas. These network structure can better complete feature matching and generate more accurate point clouds. Fig. 7 shows qualitative results of our approach and the baselines on the four scenes of Tanks and Temples benchmark.

As is visualized in Fig. 7, in large-scale and complex indoor scenes, COLMAP, CasMVSNet and TransMVSNet can basically reconstruct the geometric structure of the object completely. Compared with other methods, the generated point cloud of COLMAP, whose F-score is not the highest, is more complete but sparse. Also, its geometric structure is clearer. Yao et al. (2018) tested MVSNet on outdoor scenes in the intermediate group of Tanks and Temples and achieved good results. However, in this paper, the 3D indoor scenes reconstruction results in the advanced group are not

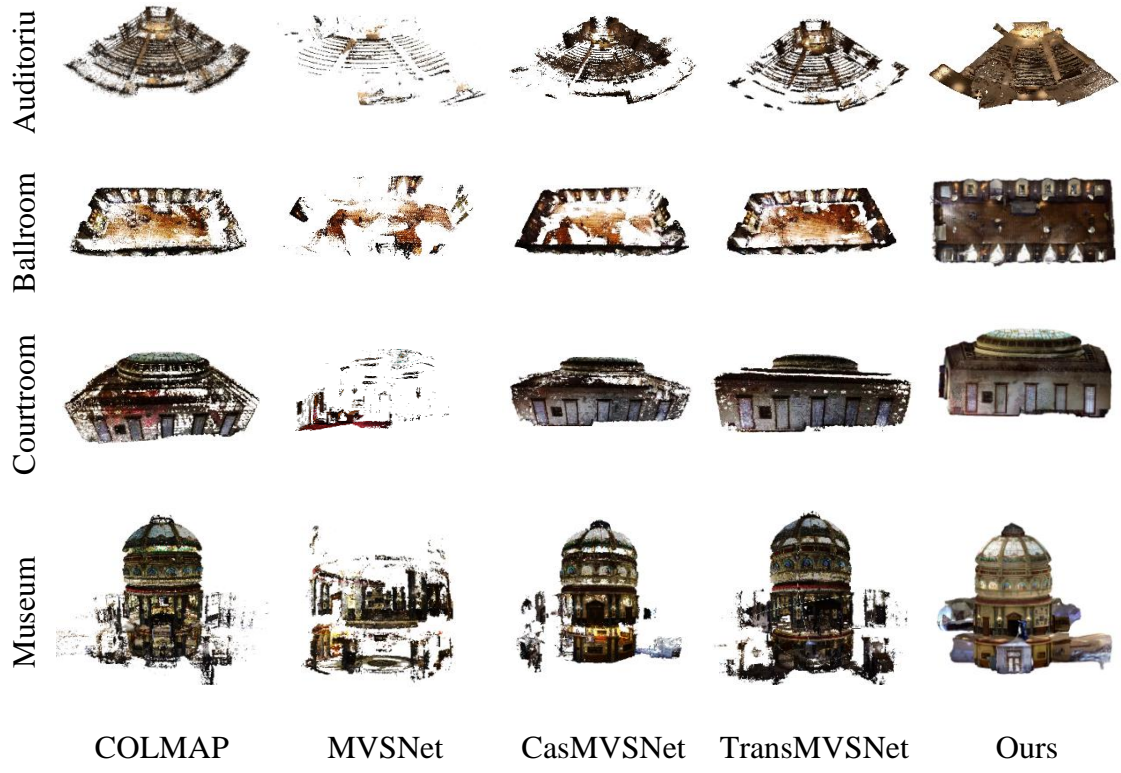

(a) Overall results of 3D reconstruction

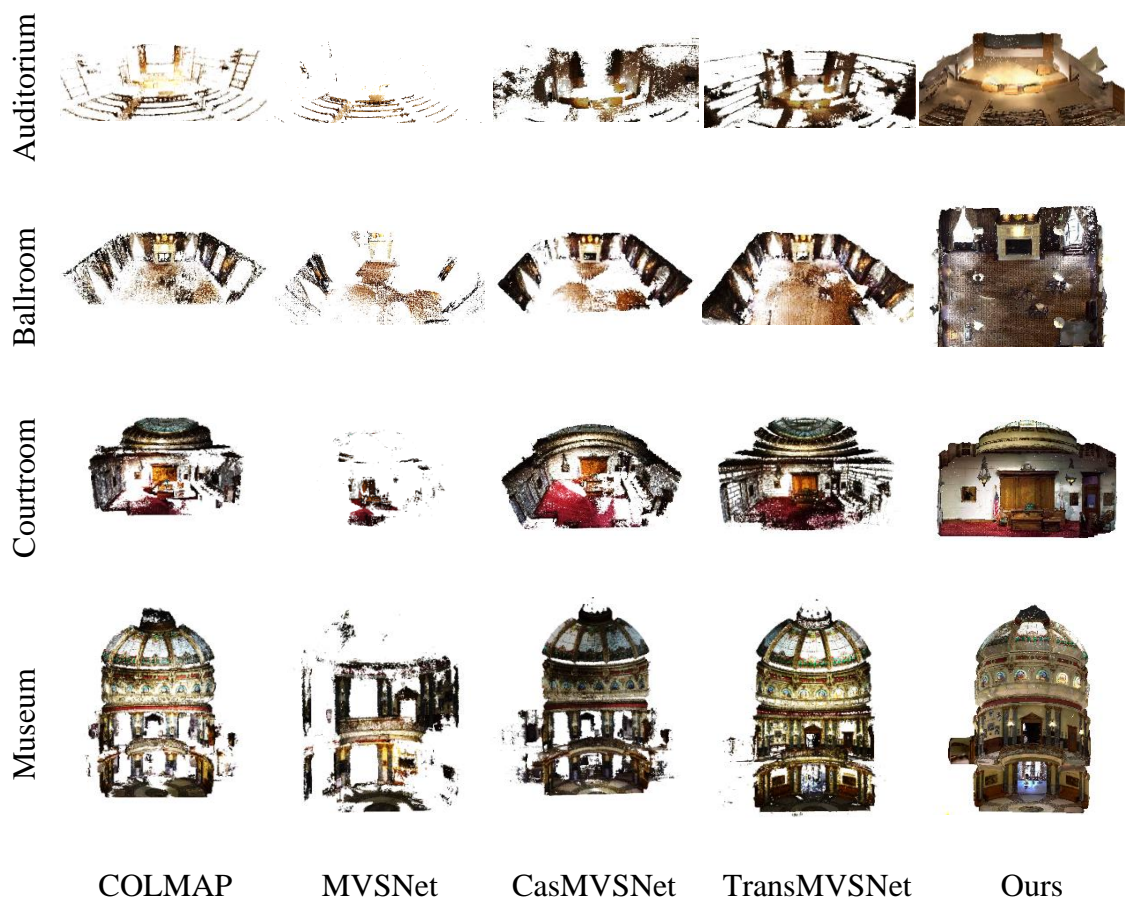

(b) Locally enlarged results of 3D reconstruction

Figure 7 Qualitative results with baselines on Tanks and Temples.

impressive, which is due to the insufficiency of MVSNet to reconstruct high-resolution scenes and the optimization of the network. CasMVSNet and TransMVSNet have significantly improved performance, and the generated point clouds are denser and more accurate. The details of the point clouds generated by CasMVSNet and TransMVSNet are obviously missing, such as the podium, curtain, furniture and wall ornaments in the scenes. In comparison, the generated point clouds of our framework are richer and clearer in detail. It can produce more reliable points on poor-texture areas, complex surfaces in Fig. 7 (b) with repetitive patterns (such as wooden tables and chairs, carpets) and reflective surfaces (glass windows and floor tiles). In short, in complex large-scale indoor scenes, ours' is not as susceptible to shooting conditions as traditional methods, and it is also more robust and reliable than other deep-learning-based 3D reconstruction methods in comparative experiments. It can generate more complete, denser, less cluttered and more detailed point clouds for high-resolution scenes under the same conditions.

**4.4 Deep-learning-based point cloud surface optimization**

In this section, the effectiveness of the proposed method is verified in the experiments of object-level and scene-level point cloud surface optimization with ShapeNet dataset (chair) and synthetic indoor scene dataset, respectively. And we compare with existing point cloud surface optimization methods.

During the training phase, we freeze MVDMM and fine-tuning the PCSOM on ShapeNet dataset and synthetic indoor scene dataset. The dataset size of the network decreases. The split of datasets follows the same setting in CONet. First, we pre-train the convolutional occupancy network with a batch size of 32 and learning rate of 0.0001 for overall 30k iterations. Then the whole network is further performed with sign-agnostic implicit surface optimization. The initial learning rate in the experiment is 0.00003, which decays by 0.3 every 400 iterations while the iteration is 1000 times. Finally, 30 scenes are randomly selected from the test set of each dataset for evaluation.

**4.4.1 Evaluation on ShapeNet dataset**

We evaluate SPSR, ONet, CONet, SA-ConvONet and our framework, i.e., SA-

ConvONet after model adaptation on ShapeNet dataset (chair) for object-level point cloud surface optimization with CD, NC, and FS as primary evaluation metrics. The results are illustrated in Tab. 4.

Table 4 Quantitative comparison of point cloud surface reconstruction on ShapeNet dataset (chair)

| Methods | CD↓ | NC↑ | FS↑ |
| --- | --- | --- | --- |
| SPSR | 1.923 | 81.54 | 80.86 |
| ONet | 1.117 | 84.58 | 62.35 |
| CONet | 0.821 | 91.12 | 74.73 |
| SA-ConvONet | 0.522 | 93.51 | 97.16 |
| MRIo3DS-Net (ours) | **0.275** | **95.67** | **98.78** |

Our framework shows an advantage in terms of visual quality in simple-structure-object-reconstruction. Compared with other methods listed in Tab. 4, Our framework shows a consistently superiority.

### 4.4.2 Evaluation on synthetic indoor scene dataset

In this section, we perform the 3D reconstruction experiment of the scene-level synthetic indoor scene dataset. The specific results are shown in Tab. 5.

Table 5 Quantitative results on synthetic indoor scene dataset

| Method | CD↓ | NC↑ | FS↑ |
| --- | --- | --- | --- |
| SPSR (Ma et al., 2020) | 2.083 | 78.21 | 76.17 |
| ONet (Mescheder et al., 2019) | 2.030 | 78.34 | 54.13 |
| CONet (Lu et al., 2019) | 2.020 | 83.43 | 73.28 |
| SA-ConvONet (Tang et al., 2021) | 0.495 | 90.04 | 93.85 |
| POCO (Boulch and Marlet, 2022) | 0.360 | 91.90 | 98.00 |
| ALTO (Wang et al., 2023b) | 0.350 | 92.10 | 98.10 |
| GridFormer (Li et al., 2024) | 0.340 | 92.60 | 98.30 |
| Sp-ConvONet (Wang et al., 2024) | 0.380 | 91.60 | 97.6 |
| MRIo3DS-Net (ours) | **0.315** | **95.40** | **98.45** |

To demonstrate the generalization ability of the method used in this paper, we perform a surface reconstruction experiment of unorientated point clouds on synthetic indoor scene dataset. Similarly, the better quantitative results in Tab. 5 also show that in indoor scenes, our framework can recover more accurate fine-grained geometries, and the effective sign-agnostic implicit surface optimization improves the universality and scalability of surface optimization for new tasks.

### 4.5 End-to-end 3D reconstruction

We computed the camera intrinsics and extrinsics using SfM in COLMAP and then processed them into the universal format of deep-learning-based MVS network. In this paper, we select four indoor objects from the optimized DTU dataset and four indoor scenes from the optimized Tanks and Temples benchmark to perform end-to-end 3D reconstruction.

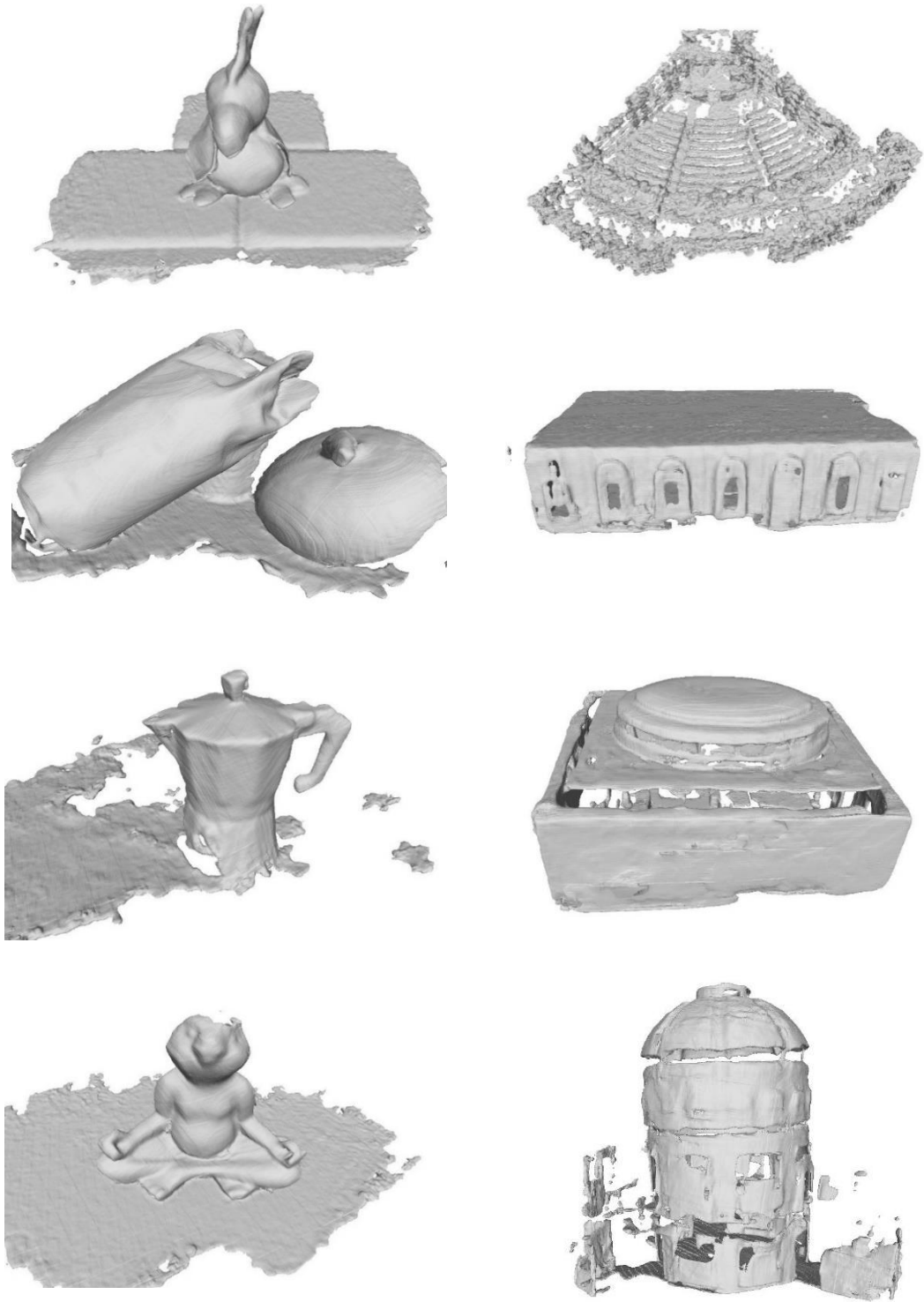

Scenes in DTU dataset　　　　　　　Scenes in Tanks and Temples benchmark

Figure 8 3D reconstruction results of MRIo3DS-Net

As is visualized in Fig. 8, the multi-task cascade network, i.e., MRIo3DS-Net framework can basically restore the main geometric structure of indoor objects or scenes, while it also suffers from surface damage and small holes in many scenes whose geometric details are not preserved well. The surface generated by MRIo3DS-Net, i.e., the point cloud surface generated by MVS, is slightly worse than the surface reconstructed by scanned point clouds, as the result of point cloud surface reconstruction is largely affected by the quality of the point cloud.

The comparison results show that the MRIo3DS-Net framework proposed in this paper can reliably and completely reconstruct the real indoor scene under a natural environment. In addition, MRIo3DS-Net can reconstruct multiple scenes while minimizing manual operations, maximize the leverage of existing equipment so that it can greatly shorten the 3D reconstruction time and improve the 3D reconstruction efficiency.

### 4.6 Multi-task loss function

We leveraged many scenes from DTU dataset and Tanks and Temples benchmark for the evaluation of designed multi-task loss function with CD, NC, FS as evaluation metrics.

The quantitative comparison results in Tab. 6 show that the two subtasks in the multi-task cascade network designed in this paper do have a certain connection. The results of MVDMM directly affect PCSOM, which improves the performance of 3D indoor scenes reconstruction to a certain extent.

Table 6 Quantitative comparison results on Tanks and Temples

| Method | CD↓ | NC↑ | FS↑ |
|---|---|---|---|
| SA-ConvONet (Tang et al., 2021) | 0.515 | 89.35 | 92.70 |
| MRIo3DS-Net (ours) | **0.242** | **94.28** | **98.60** |

As shown in Fig. 9, after automatically adjusting and optimizing the weights of the subtasks, the 3D reconstruction results of MRIo3DS-Net are generally better than that of SA-ConvONet. The 3D reconstruction results of MRIo3DS-Net are more complete but rough. For example, the food packaging bag in the second scene is more complete, while the third and fourth scenes are more complete but have a rougher surface, which means that the density of the generated point cloud increased while somewhat added some noise.

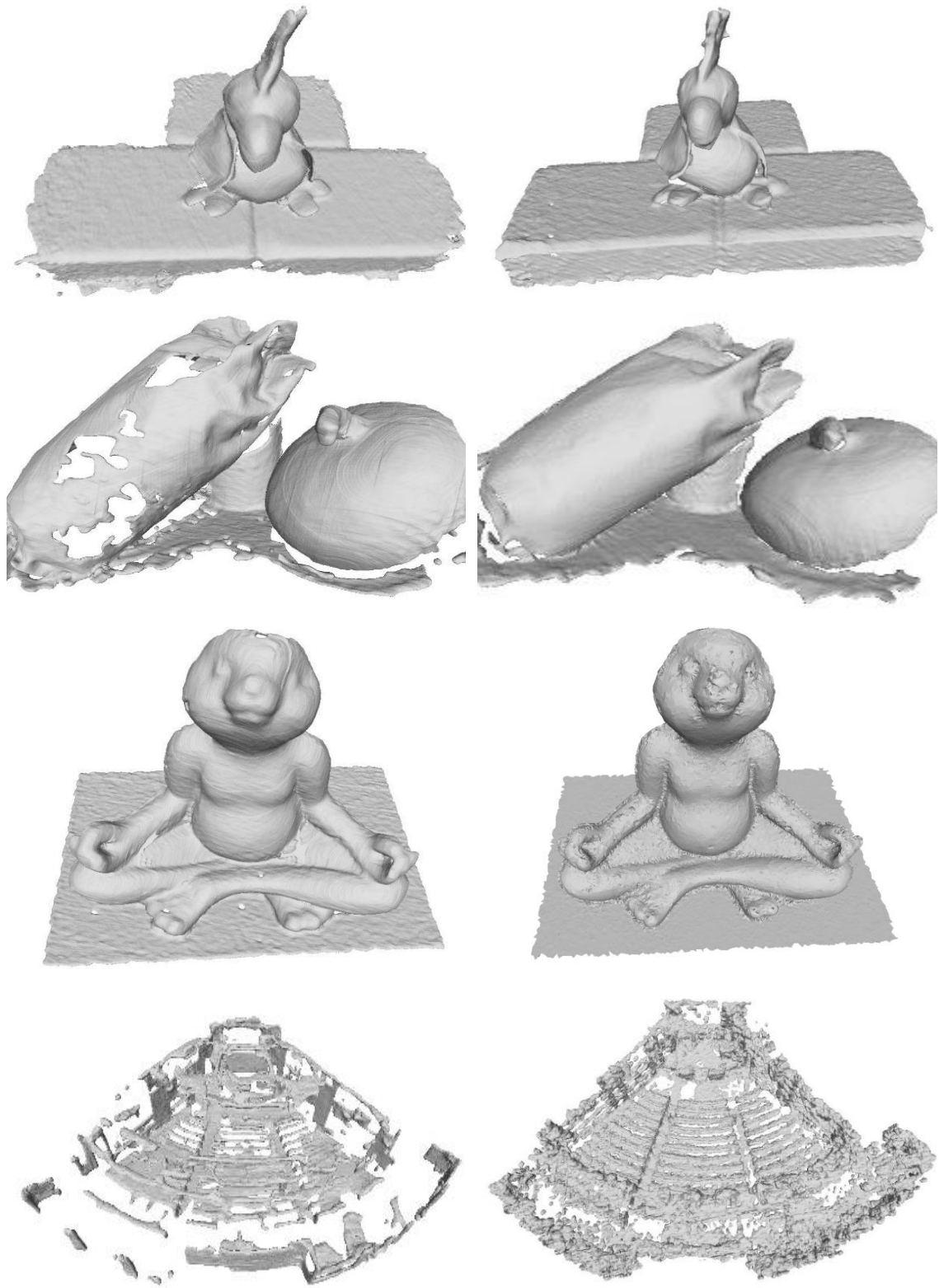

(a)SA-ConvONet (b)MRIo3DS-Net

Figure 9 Comparison between SA-ConvONet and MRIo3DS-Net

### 4.7 Ablation studies

In this section, we perform ablation studies and quantitative analysis on Tanks and Temples benchmark to evaluate the effectiveness of different modules in MRIo3DS-Net framework proposed in this paper with CD, NC and FS as evaluation metric (Tab. 7).

Table 7 Quantitative performance with different components on Tanks and Temples.

|  | Model Settings | | | | CD↓ | NC↑ | FS↑ |
| --- | --- | --- | --- | --- | --- | --- | --- |
|  | MVDMM | PCSOM | Smooth loss | RNN-like reinforcement | | | |
| (a) | × | √ | × | × | - | - | - |
| (b) | √ | √ | × | × | 0.515 | 89.35 | 92.73 |
| (c) | × | √ | √ | × | - | - | - |
| (d) | √ | √ | √ | × | 0.451 | 92.24 | 96.05 |
| (e) | √ | √ | √ | √ | **0.242** | **94.28** | **98.60** |

Due to lack of multi-view dense matching results, we did not provide the accuracy for experiments (a) and (c). After applying the smooth loss, there is a significant decrease in CD, accompanied by an increase in NC and FS, indicating the effectiveness of the smooth loss. The complete MRIo3DS-Net framework achieves SOTA performance.

The ablation experiment results show that the MRIo3DS-Net framework, which integrates MVDMM and PCSOM, achieves the result of "1+1>2" compared with the single task, which is because MRIo3DS-Net framework fuses the two modules together so that creating a synergy that makes the overall 3D reconstruction far more effective than when the components are used alone. Through this integration, it is not simply to fuse the two modules together, but to promote and complement each other, resulting in a more powerful and efficient output.

## 5 Discussion

We propose an end-to-end 3D reconstruction framework and design a multi-task loss function based on Bayesian uncertainty for this framework. Extensive experiments have demonstrated that the MRIo3DS-Net framework produces superior 3D reconstruction results compared to traditional algorithms. It can reliably and completely reconstruct the real indoor scene under the natural environment, which has certain potential for the 3D reconstruction of new target domain task, and greatly reduces the running time of 3D reconstruction and improves the 3D reconstruction efficiency.

This section discusses the factors that impact the quality of reconstruction and the generalization ability of 3D reconstruction using deep learning.

## 5.1 Factors affecting the 3D reconstruction quality

### 5.1.1 Multi-view dense matching networks

In the multi-view dense matching task, due to the greatest influence on 3D reconstruction, image feature extraction, multi-view image matching and loss function are the most improved modules. Among the four networks, i.e., COLMAP, MVSNet, CasMVSNet and TransMVSNet evaluated in this paper, TransMVSNet achieves remarkable performance. The COLMAP, whose 3D reconstruction results in poor texture or non-Lambertian surfaces are incomplete and the point clouds are sparse, applies scale invariant feature transform for feature detection and similarity measurement for feature matching. The 3D reconstruction results obtained by MVSNet using 2D convolutional neural network to learn global semantic information are more robust, but due to the limitation of memory, they cannot be used well for large-scale 3D reconstruction of high-resolution images. CasMVSNet applies the cascade cost volume and feature pyramid to extract the multi-scale feature map to complete the depth estimation based on the coarse-to-fine regularization pattern, and the generated point cloud is denser. Based on CasMVSNet, TransMVSNet introduces Transformer to aggregate global and local context information. In addition, it uses focal loss to strengthen supervision of ambiguous areas so that it can optimize depth estimation, further to obtain more accurate 3D reconstruction results.

### 5.1.2 Point cloud surface optimization networks

SA-ConvONet can reconstruct the surface of scanning point clouds without normal orientation or point clouds generated by multi-view dense matching networks. After pre-training on the dataset with ground truth surface, it provides the initial neural implicit field for the test phase, so that the network can further optimize the implicit field network by utilizing the unsigned cross-entropy loss without using normal orientation in the test phase. SPAR completes pint cloud surface reconstruction through surface normal estimation, but it also accumulates a lot of errors, which lead to defects in the reconstructed surface mesh. Compared with ONet that only use global shape features, SA-ConvONet uses 3D U-Net to aggregate global and local shape features, which can not only recover the surface details of the target object, but also enforce the geometric consistency between learned local geometries. In addition, SA-ConvONet fine-tuned the prior shape information obtained from the pre-trained model during the test phase, so it has an advantage over ONet and CONet in reconstructing the details of the surface model, such as chair legs, lamp posts, etc.

### 5.1.3 Multi-task loss function

Many deep learning applications benefit from multi-task learning. Compared to learning tasks in isolation, multi-task learning achieves better performance, computational efficiency, and memory usage (Vandenhende et al., 2021). The performance of multi-task learning heavily depends on the relative weights of each task's loss. Manually tuning these weights is considered a challenging and costly process (Kendall et al., 2018). In this paper, we employ a multi-task loss function that balances multiple loss functions by considering the mean squared error uncertainty of each task. This enables the simultaneous acquisition of diverse quantities across multiple tasks, encompassing various units and scales (Kendall et al., 2018).

In this paper, binary cross-entropy loss, focal loss, and smooth loss are applied to the multi-task learning, with their weights balanced using a multi-task loss function. From the ablation experiments, the smooth loss demonstrates significant effectiveness. By imposing smooth constraints, we can enhance the accuracy of 3D surfaces, thereby guiding multi-view dense matching to achieve higher precision. This process provides high-quality data sources for point cloud surface optimization, creating a mutually beneficial cycle of improvement (Vu et al., 2011).

### 5.1.4 End-to-end 3D reconstruction RNN-like network framework

The proposed MRIo3DS-Net framework does not require intervention during runtime and runs faster than two separate networks. The framework consists of two mutually reinforcing modules that are scalable and can be replaced by higher performance networks of the same type.

Furthermore, the MRIo3DS-Net framework optimizes the two modules separately before cascading them. The designed multi-task loss function optimizes the weights for sub-tasks, resulting in superior accuracy and completeness of the reconstructed 3D model compared to a single multi-view dense matching network and a point cloud surface optimization network. The innovation of the end-to-end 3D reconstruction framework lies in its fusion of two distinct modules, combining the strengths of each component, resulting in 3D reconstruction results that far exceed what can be achieved when using them independently. Specifically, improvements in MVDMM can provide more accurate inputs to PCSOM. The PCSOM provides feedback, i.e., finely optimized 3D surface and backpropagation of the deep neural network, to recursively reinforce the differentiable warping for optimizing MVDMM, aiming to achieve more accurate matching results and obtain a more precise point cloud. This forms a RNN-like optimization strategy, creating a synergistic effect that leads to an overall improved performance (Choy et al., 2016; Liu and Ji, 2020). Additionally, within PCSOM, smooth loss is applied to impose smooth constraints on the reconstructed point cloud surface, resulting in a more accurate 3D surface (Vu et al., 2011). This reduces warping errors caused by inaccuracies in the 3D surface and improves the accuracy of 3D reconstruction.

**5.2 Generalization ability of deep-learning-based 3D reconstruction**

Deep-learning-based 3D reconstruction methods often perform well in many scenes but may encounter some challenges when applied to new datasets. The MRIo3DS-Net framework addresses this issue by utilizing a model-adaptation strategy to fine-tune MVDMM and PCSOM, offering three advantages: (1) Address the issue of poor generalization in 3D reconstruction based on DNNs to some extent. Deep-learning-based 3D reconstruction is more universal; (2) Accelerate the convergence of the two modules, MVDMM and PCSOM; (3) Make the two modules within the MRIo3DS-Net framework to mutually promote and iteratively evolve over time. Therefore, the MRIo3DS-Net framework has the potential to be effective for reconstructing new target domain tasks.

# 6 Conclusion

Based on existing 3D reconstruction methods, focusing on indoor scenes, this paper primarily investigates two crucial aspects of the 3D reconstruction: multi-view dense matching and point cloud surface optimization and proposes an end-to-end mutually reinforcing RNN-like framework for 3D indoor scenes reconstruction based on DNNs. Additionally, we design a multi-task loss function based on Bayesian uncertainty for this framework. The network cascades two mutually reinforcing modules, i.e., multi-view dense matching and point cloud surface optimization, to reconstruct detailed and complete indoor scenes. It addresses the inefficiency of traditional methods and fills the gap of incomplete flow in learning-based 3D reconstruction. Additionally, it enhances the automation, intelligence, and integration levels of 3D reconstruction. Moreover, it has certain potential for the new target domain task.

However, this paper still has some limitations. Firstly, the reconstruction integrity of edges in the model still needs improvement. The point cloud generated after depth map fusion requires further noise filtering to ensure point cloud integrity while enhancing accuracy. Secondly, we consider the reasonable densification of sparse point clouds or the rational filtering and down-sampling of dense point clouds to improve the quality of point clouds.

**Acknowledgements**

The authors are grateful for the comments and contributions of the editors, anonymous reviewers and the members of the editorial team. This work was supported by the Key Program of the National Natural Science Foundation of China under Grant Nos. 42030102, National Natural Science Foundation of China (NSFC) under Grant Nos. 42201474, 41771493 and 41101407, and the Fundamental Research Funds for the Central Universities under Grant CCNU22QN019.

**Declaration of competing interest**

The authors declare that they have no known competing financial interests or personal relationships that could have appeared to influence the work reported in this paper.